\documentclass[twocolumn]{IEEEtran}
\usepackage[dvipsnames]{xcolor}
\usepackage{cite} 
\usepackage[pdftex]{graphicx} 
\usepackage{amsmath} 
\usepackage{eso-pic,graphicx}
\usepackage{algorithmic}
\usepackage{array} 
\usepackage[caption=false,font=footnotesize]{subfig} 
\usepackage{dblfloatfix}
\usepackage{url} 

\usepackage{graphicx}   
\usepackage{adjustbox}  
\usepackage{float}      
\usepackage{microtype}
\usepackage[utf8]{inputenc} 
\usepackage[T1]{fontenc}    
\usepackage{paralist} 
\usepackage{marvosym} 
\usepackage{bm} 
\usepackage{bbm} 
\usepackage{amssymb}   
\usepackage{mathrsfs} 
\usepackage{tabularray} 
\UseTblrLibrary{booktabs}       
\usepackage{tabularx} 
\usepackage[flushleft]{threeparttable} 
\usepackage{colortbl} 
\usepackage{makecell} 
\usepackage{multirow} 
\usepackage{nicefrac} 
\usepackage[dvipsnames]{xcolor} 
\usepackage{soul} 
\setul{0.5ex}{0.3ex}

\definecolor{citecolor}{RGB}{34,139,34}
\usepackage[pagebackref=true,breaklinks=true,letterpaper=true,colorlinks,
    citecolor=citecolor,bookmarks=false]{hyperref}
\usepackage{cleveref}  
\crefformat{figure}{Fig.~#2#1#3}
\crefformat{equation}{Eq.~(#2#1#3)}
\crefformat{table}{Table~#2#1#3}
\crefformat{section}{Section~#2#1#3}
\crefformat{algorithm}{Algorithm~#2#1#3}
\crefformat{appendix}{Appendix #2#1#3}
\crefformat{subsection}{subsection~#2#1#3}
\usepackage{cellspace}
\usepackage[numbers,sort&compress]{natbib} 

\usepackage{xspace} 

\usepackage{pifont} 
\definecolor{Gray}{rgb}{0.9,0.9,0.9}
\definecolor{LightCyan}{rgb}{0.88,1,1}
\newcolumntype{a}{>{\columncolor{Gray}}c}
\newcolumntype{b}{>{\columncolor{white}}c}

\begin{document}

\setlength{\abovedisplayskip}{.5\baselineskip} 
\setlength{\belowdisplayskip}{.5\baselineskip} 

\title{DSCSNet: A Dynamic Sparse Compression Sensing Network for Closely-Spaced Infrared \\ Small Target Unmixing}



\author{
  Zhiyang~Tang$\dagger$,
  Yiming~Zhu$\dagger$, 
  Ruimin Huang,
  Meng Yang,
  Yong Ma,
  Jun Huang,
  Fan Fan*
  
  \thanks{
    This research was supported by the National Key R$\&$D Program of China under Grant 2024YFE0202500, 
National Natural Science Foundation of China (No. 62475199, 62473297, and U23B2050)\\

    \emph{The first two authors $\dagger$ contributed equally to this work. (Corresponding author: Fan Fan)}
    }

}

\maketitle

\begin{abstract}
Due to the limitations of optical lens focal length and detector resolution, distant clustered infrared small targets often appear as mixed spots. The Close Small Object Unmixing (CSOU) task aims to recover the number, sub-pixel positions, and radiant intensities of individual targets from these spots, which is a highly ill-posed inverse problem. Existing methods struggle to balance the rigorous sparsity guarantees of model-driven approaches and the dynamic scene adaptability of data-driven methods. To address this dilemma, this paper proposes a Dynamic Sparse Compressed Sensing Network (DSCSNet), a deep-unfolded network that couples the Alternating Direction Method of Multipliers (ADMM) with learnable parameters. Specifically, we embed a strict $\ell_1$-norm sparsity constraint into the auxiliary variable update step of ADMM to replace the traditional $\ell_2$-norm smoothness-promoting terms, which effectively preserves the discrete energy peaks of small targets. We also integrate a self-attention-based dynamic thresholding mechanism into the reconstruction stage, which adaptively adjusts the sparsification intensity using the sparsity-enhanced information from the iterative process. These modules are jointly optimized end-to-end across the three iterative steps of ADMM. Retaining the physical logic of compressed sensing, DSCSNet achieves robust sparsity induction and scene adaptability, thus enhancing the unmixing accuracy and generalization in complex infrared scenarios. Extensive experiments on the synthetic infrared dataset CSIST-100K demonstrate that DSCSNet outperforms state-of-the-art methods in key metrics such as CSO-mAP and sub-pixel localization error.

\end{abstract}

\begin{IEEEkeywords}
Clustered small targets; infrared imaging; target unmixing; compressed sensing; deep-unfolded network; ADMM; dynamic thresholding; sparsity constraint
\end{IEEEkeywords}
\vspace{-1\baselineskip}

\section{Introduction} 
\label{sec:introduction}

Infrared imaging technology captures the thermal radiation from objects, enabling an inherent all-weather imaging capability \cite{deng2016small, zhu2025shifting}. It is widely applied in remote sensing, security monitoring, and surveillance \cite{zhu2025toward}. In these scenarios, targets of interest are often imaged as small targets due to long detection distances, making infrared small target detection (IRSTD) a research field with significant practical applications\cite{wu2022srcanet}. However, with the development of swarm control technologies for low-altitude unmanned aerial vehicles (UAVs)\cite{xiao2024background}, the accurate interpretation of clustered infrared small targets has become a critical demand in national defense monitoring and low-altitude security \cite{zhao2022single}.

\begin{figure}
    \centering
    \includegraphics[width=1\linewidth]{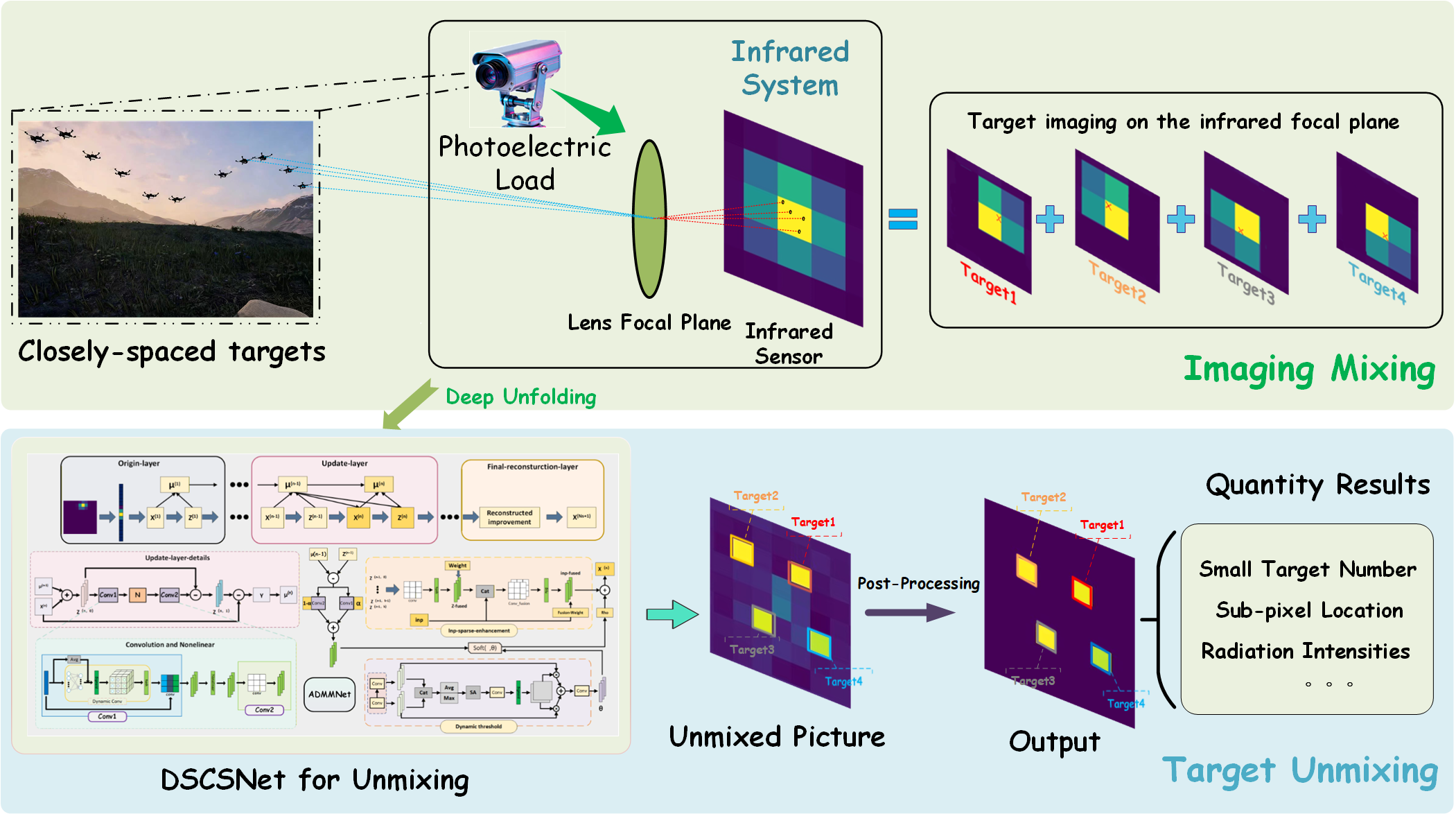}
    \caption{The process of imaging mixing and unmixing of Closed-spaced targets. The green box represents the process of mixing distant neighboring targets in the infrared system, menawhile, the blue box represents the process of unmixing. The input for unmixing is the mixed image, and the output is the number of small targets in the mixed light spot, as well as the sub-pixel position and radiation intensity of each small target.}
    \label{fig:fig1_unmixing_process}
\end{figure}

\begin{figure}
    \centering
    \includegraphics[width= 1.0\linewidth]{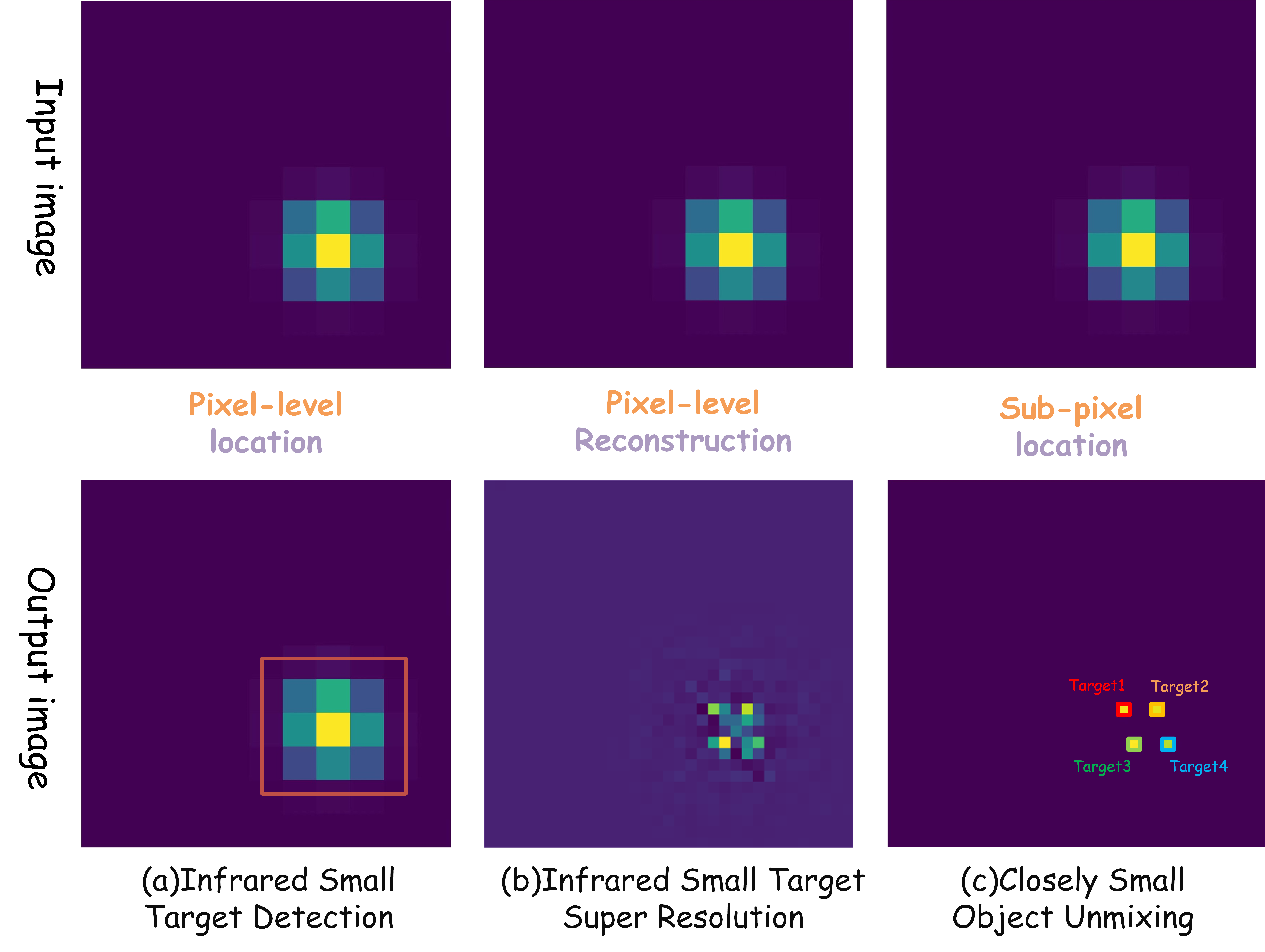}
    \caption{The first column shows that IRSTD methods assume a one-to-one correspondence between detected targets and real-world objects; when only one spot appears in the image, the detection result corresponds to a single target. The super-resolution approach in the second column can split the mixed spot into several sub-targets, yet its sub-pixel localization accuracy is insufficient, leading to missed detections and false alarms. In contrast, the DCSCNet unmixing method in the third column delivers more refined detection, enabling sub-pixel localization within the spot and unmixing of potential sub-targets.}
    \label{fig:fig2diff_compare}
\end{figure}

A particularly challenging scenario arises in clustered target detection: due to the long imaging distance, the angular separation between clustered targets often falls below the optical diffraction limit (Rayleigh criterion) of the infrared system. As shown in Fig. \ref{fig:fig1_unmixing_process}, their Point Spread Functions (PSFs) severely overlap on the focal plane, convolving into a single indistinguishable mixed spot \cite{wang2024improving}. Consequently, this phenomenon is well beyond the scope of traditional tasks such as infrared small target detection (IRSTD) \cite{dai2021attentional, li2022dense, wu2022uiu, zhu2024toward, jing2025swan} and image super-resolution (SR) \cite{dong2015image, wang2018esrgan, zhang2021designing}, giving rise to a core task termed "\textbf{Close Small Object Unmixing} (CSOU)". The core of CSOU is to recover the intrinsic attributes of each constituent target from the mixed spot, which necessitates the simultaneous achievement of three objectives:
\begin{enumerate}
    \item \textbf{Sub-pixel Target Detection:} As illustrated in \textbf{Fig. \ref{fig:fig2diff_compare}}, unlike IRSTD that focuses on pixel-level target detection, CSOU demands accurate estimate of the number of targets within the mixed light spot rather than mere target-level detection.
    \item \textbf{Sub-pixel Localization:} Distinct from IRSTD which realizes pixel-level target localization, CSOU needs to determine the sub-pixel coordinates of each individual target.
    \item \textbf{Recovery of intrinsic target information:} Distinct from image SR that aims to improve the overall image resolution, CSOU needs to invert both the sub-pixel coordinates and radiant intensity of each target.
\end{enumerate}

    This is different from the traditional IRSTD methods (such as GSTUNet, DNANet, and FDA-IRSTD)\cite{zhu2024toward, li2022dense, zhu2025toward}, which focus on pixel-level object region segmentation for small object detection, as shown in Fig. \ref{fig:fig2diff_compare}(a). These methods rely on a one-to-one correspondence between the object to be detected and the annotation mask. As a result, they completely ignore the need for cross-scene Target Unmixing (CSOU) to resolve the intrinsic properties of multiple targets from mixed signals, such as target count, sub-pixel position, and radiance intensity.
The inherent differences in core objectives mean that CSOU is a novel and challenging problem distinct from traditional tasks.

%
From a signal processing perspective, CSOU can be naturally formulated within the framework of compressed sensing (CS) and reconstruction \cite{donoho2006compressed}. It aims to reconstruct a high-resolution target distribution from low-resolution observations. The formulation is validated by two physical priors. First, clustered infrared small targets are spatially sparse in the entire imaging plane, i.e., they only occupy an extremely small part of the space (most areas have no targets). \cite{lin2011resolution}; Second, "long-distance imaging" will cause the "point spread Functions (PSFs)" of multiple infrared small targets to overlap with each other, and the respective infrared energies of the targets will mix together, eventually forming "mixed spots" on the imaging plane. We can only observe the pixel brightness of the mixed spots and cannot directly distinguish the brightness and position of individual targets. \cite{hui2013super}. The goal of CSOU is to utilize mixed spots to restore sparse attributes such as counting, sub-pixel coordinates, and radiation intensity - this is in perfect alignment with the theory of compressed sensing, making this framework mathematically rigorous and grounded in physics.


Guided by the theory of compressed sensing, existing CSOU approaches can be broadly categorized into two paradigms: traditional model-driven methods \cite{Daubechies2004ISTA} and deep learning-based data-driven methods \cite{zhang2018ista}.
Specifically, the traditional model-based approaches are to establish inverse problem optimization models related to data fidelity and sparse regularization, and then solve them through hand-crafted priors and fixed regularization hyperparameters, such as the Iterative Contraction Threshold Algorithm (ISTA) \cite{Daubechies2004ISTA} and ADMM \cite{boyd2011distributed} Here, the hand-crafted priors refer to manually designed prior constraints, and regularization hyperparameters are fixed parameters used to adjust the balance between model complexity and data fitting.

While these model-driven approaches provide rigorous mathematical interpretability and are suitable for simple scenarios \cite{yang2018admm}, they face significant challenges in CSOU tasks: fixed parameters are difficult to adapt to scenarios with different number of targets, locations, and dynamically varying radiation intensities. In particular, pre-defined regularization hyperparameters fail to adaptively balance the sparsity constraints with the data fidelity. Consequently, these methods suffer from poor generalization performance and require tedious hyperparameter tuning, which renders them ineffective for complex, dynamic infrared scenarios \cite{zhang2018ista}.

Data-driven deep learning approaches can generally be categorized into two groups: super-resolution (SR)-based methods and deep unfolding–based methods. The image super-resolution algorithms improve the overall clarity of targets by enhancing the perceptual quality of individual mixed spots \cite{dong2015image, wang2018esrgan, zhang2021designing}, thereby enabling target detection methods to detect sub-targets. However, these approaches primarily emphasize visual fidelity rather than faithful physical signal recovery and therefore struggle to effectively disentangle the coupled information of closely spaced sub-targets.  
To overcome the above drawback, the deep unfolding-based methods leverage model-driven optimization frameworks and embed them into learnable networks. Representative approaches, such as ISTA-Net \cite{zhang2018ista}, ADMM-Net \cite{yang2018admm} and DISTANet \cite{han2025dista}, incorporate physical and hand-crafted priors into data-driven architectures to enable more reliable recovery of sub-targets. Nevertheless, two key limitations remain:
\begin{enumerate}
    \item \textbf{Insufficient robustness in complex infrared scenes:} As the complexity of mixed spot signal patterns increases—such as when the number of densely distributed targets grows—accurate reconstruction of a small target to a single pixel becomes challenging, often manifesting as varying degrees of artifacts around a given pixel. This leads to a decline in network prediction accuracy and a corresponding degradation in reconstructed image quality.
    \item \textbf{Insufficient adaptation of network sparsity:} The CS mechanisms embedded in these frameworks often promote excessive image smoothing and fail to impose sparsity constraints tailored to small infrared targets, thereby limiting further improvements in unmixing accuracy.
\end{enumerate}



To address the aforementioned limitations, we revisit the CSOU problem from a prior-driven perspective. As a highly ill-posed inverse problem\cite{han2025dista}, the successful resolution of CSOU fundamentally depends on the accurate exploitation of prior knowledge regarding target radiation signals. In this context, we argue that an ideal CSOU framework should possess two key capabilities:
\begin{enumerate}    
    \item \textbf{Dynamic Scene-Adaptive Perception:} In real-world scenarios, target states and interference conditions evolve dynamically. When small targets are densely distributed, their local spatial proportion becomes significant, necessitating stronger sparsity constraints (e.g., through larger penalty coefficients or stricter thresholds). These constraints help the model focus on selecting only the most salient target features, thereby preventing it from being distracted by excessive background details or noise. Conversely, in low-density target scenarios, weaker sparsity constraints are required, allowing the model to flexibly activate a broader set of features to capture potential target signals without overly suppressing the background. Therefore, an effective model must possess the ability to intelligently perceive the contextual dynamics and adjust sparsity constraints accordingly.
    \item \textbf{Strong Sparsity Induction}: The observed mixed signals typically originate from a limited number of high-energy point sources, implying that the underlying target distribution is inherently sparse. Accordingly, the unmixing process should avoid indiscriminate spatial smoothing and instead emphasize the preservation of both the radiation intensity and spatial location of each individual sub-target during signal recovery, thereby enabling the clear separation of overlapping or closely spaced targets.

\end{enumerate}



The core motivation of this work arises from the failure of existing unmixing frameworks to reconcile the rigorous sparsity guaranties of model-based methods with the dynamic adaptability of data-driven approaches\cite{monga2021algorithm}. We propose that the key lies in creating a novel architecture where sparsity constraints evolve from static, manually defined rules into intelligent, context-aware processes tailored to specific input signals\cite{shen2022transcs}. This architecture is realized by advancing the deep unfolding paradigm \cite{mansouri2024robust}, specifically through a dynamically parameterized ADMM framework that decouples sparsity regularization from data fidelity, thereby establishing the theoretical groundwork for our proposed DCSCNet.
\begin{figure}
    \centering
    \includegraphics[width=1\linewidth]{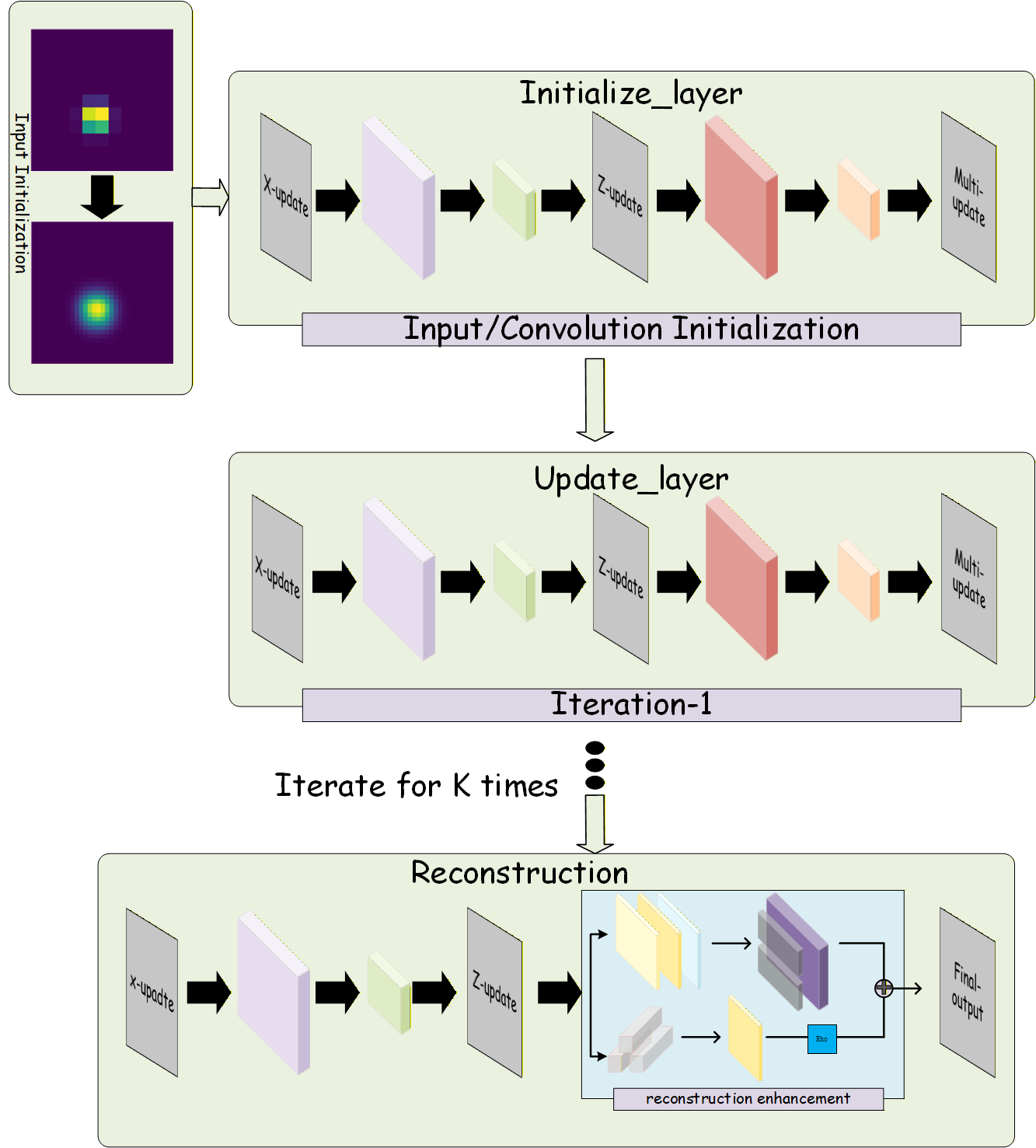}
    \caption{Schematic of DCSCNet unmixing, organized into three main stages—initialize\_layer, update\_layer, and reconstruction\_layer. Within each stage, three embedded sub-layers perform the ADMM variable updates for input image X, auxiliary variable Z, and multiplier u.}
    \label{fig:placeholder}
\end{figure}



Based on the above analysis, we propose a novel \textbf{Dynamic Sparse Compressed Sensing Network} (DSCSNet). Built upon the deep unfolding theory, DSCSNet deeply couples the physically interpretable ADMM optimization framework with dynamic, data-driven learning.

We propose three key innovations. \textbf{First}, end-to-end optimization across ADMM steps (Initialize, Update, Reconstruction, \textbf{Fig.~3}) ensures adaptive sparse iteration of parameters $x$, $\mathbf{z}$, $u$: dynamic convolution and self-attention enable threshold sparsity in $\mathbf{z}$-update; $\ell_1$-norm and sparse $\mathbf{z}$ separate targets in $x$-update; step sizes are learned in $u$-update. This preserves compressed sensing physics while enabling sparsity and scene adaptation.
\textbf{Second}, a Dynamic Threshold Generator (DTG) distills global context to predict spatially-varying thresholds per iteration, enabling content-aware feature selection.
\textbf{Third}, we replace smoothness with a strict $\ell_1$-norm constraint in $\mathbf{z}$-update and design a Dynamic Information Reorganization (DIR) module to reintroduce sparse auxiliary information in reconstruction, aligning optimization with target sparsity and avoiding smoothing bias.

The main contributions of this paper are summarized as follows:
\begin{enumerate}
\item \textbf{Theoretical Innovation: A Paradigm Fusing Physical Priors with Data-Driven Learning:} We establish a novel hybrid paradigm that integrates the sparse priors of model-based compressed sensing with the adaptive learning capabilities of deep learning. This approach overcomes the limitations of traditional methods that rely on manually preset sparse constraints and purely data-driven methods.
\item \textbf{Architectural Innovation: A Task-Specific Deep Unfolding Framework:} We design a deep unfolding architecture based on the ADMM optimization algorithm that is specifically tailored for the CSOU. This framework resolves the task mismatch bottleneck by ensuring a principled alignment between the unmixing problem's structure and the network's design.
\item \textbf{Methodological Innovation: Effective combination of dynamic threshold processing and sparse iterative information:} We deeply embed the strict $\ell_1$-norm sparsity constraint into the network's iterative $\mathbf{z}$ update stage, and in the final reconstruction stage, we integrate the iterative information and apply a joint dynamic threshold mechanism to perform adaptive filtering on the input based on the input-generated threshold. These mechanisms are jointly optimized in an end-to-end manner, enabling the precise separation of closely spaced targets in dynamic high-density scenarios.
\end{enumerate}


\section{Related Work} \label{sec:related}

\subsection{Infrared Small Target Detection}

Infrared Small Target Detection (IRSTD) aims to identify pixel-level targets against complex backgrounds, and research in this field has evolved from handcrafted feature-based methods to deep learning paradigms. Early methods relied on local contrast mechanisms \cite{chen2013local} and low-rank decomposition \cite{feng2023coarse}, but these methods struggled with poor generalization ability. The emergence of datasets like SIRST \cite{dai2021attentional} and IRSTD-1k \cite{zhang2022isnet} has catalyzed the development of data-driven approaches. Data-driven methods, especially deep learning ones, have triggered a paradigm shift in IRSTD via end-to-end feature learning, achieving remarkable performance gains on multiple public datasets. They follow several key technical evolution paths:
\begin{enumerate}
\item \textbf{Multi-scale Feature Fusion} has become the mainstream strategy to address intrinsic feature sparsity \cite{zhou2025hmfenet}. DNANet \cite{li2022dense} employs dense nested connections with gradient routing to preserve target details across scales. UIU-Net \cite{wu2022uiu} introduces a U-Net within U-Net architecture with dual attention gates for precise localization. MTMLNet \cite{MTMLNet} designs multi-stage feature aggregation and hierarchical feature fusion to capture features with different receptive fields. WMRNet \cite{WMRNet} employs discrete wavelet transform to decompose images into subbands, minimizing frequency interference in state space models.  ACMNet \cite{dai2024pick} proposes asymmetric contextual modulation that bridges semantic gaps via point-wise channel attention, achieving a 0.95 F1-measure on IRSTD-1k.
\item \textbf{Attention Mechanisms} have further enhanced robustness against background clutter. RLGA-Net \cite{wang2024improving} integrates reinforcement learning with global context boundary attention, reducing false alarms caused by bright noise. OSFormer \cite{OSFormer} introduces Varied-Size Patch Attention and Doppler Adaptive Filter for background suppression in video sequences. Lin et al. \cite{Lin2024} propose contrast-enhanced shape-biased representations via cascaded contrast-shape encoder with attention guidance. LoGoNet \cite{li2023logonet} adopts local-to-global fusion with deformable convolutions, improving the detection rate in cluttered scenes.
\item \textbf{Transformers for Global Dependency Modeling}: Recently, Transformers have been introduced to IRSTD for global dependency modeling (addressing CNNS’ receptive field limitations) \cite{zhang2026fsatfusion, hu2025datransnet}: DATransNet \cite{hu2025datransnet} employs a Spatial-Channel Cross Transformer to model long-range spatial-channel dependencies for accurate target context extraction; DSTransNet \cite{huang2025Dstransnet} and its variants balance efficiency and global modeling via hierarchical structures and shifted window attention, enabling the processing of high-resolution infrared images.
\item \textbf{Joint Learning Frameworks}: Joint optimization of multiple tasks provides complementary supervision for infrared small target detection. SRS \cite{SRS} presents a Siamese Reconstruction-Segmentation Network with dynamic-parameter convolution, using reconstruction as an auxiliary task. MTMLNet \cite{MTMLNet} further explores multi-task mutual learning to leverage both detection and segmentation annotations. 
\end{enumerate}

We emphasize that IRSTD primarily identifies whether a pixel region contains a target. However, CSOU aims to inversely resolve the exact number, sub-pixel coordinates, and radiation intensities from a single blended spot—a fundamental difference in task objectives. Consequently, existing detection methods fail to meet the core demands of CSOU, such as the requirements of sparse signal reconstruction and scene adaptivity, as shown in Fig.~\ref{fig:fig2diff_compare}. IRSTD typically outputs binary masks indicating target presence, while CSOU requires resolving overlapping signatures into discrete targets with sub-pixel-level precision. This limitation stems from their inherent architectural focus on classification rather than inverse problem solving.

\subsection{Deep Unfolding Network Learning}
With the development of deep learning, deep unfolding bridges model-based optimization and data-driven learning by translating iterative algorithms into network layers \cite{zhang2025exploring}. This constructs an end-to-end model that enjoys both physical interpretability and data-adaptive learning capabilities, achieving breakthroughs in sparse signal processing tasks such as compressive sensing and image reconstruction. For instance, \textbf{ISTA-based unfolding} initiated this paradigm: LISTA\cite{gregor2010learning} first unfolded ISTA into feedforward networks, while ISTA-Net\cite{zhang2018ista} incorporated convolutional transforms, improving PSNR over traditional CS. ISTA-Net++\cite{you2021ista} introduced adaptive thresholding across layers. Meanwhile, \textbf{ADMM-based frameworks} offered enhanced stability through variable splitting. ADMM-Net\cite{yang2018admm} unfolded the ADMM algorithm with learned parameters, reducing MRI reconstruction time while maintaining image quality. HQS-Net\cite{liu2019alista} employed half-quadratic splitting with deep denoiser priors, demonstrating superior performance in Poisson noise conditions.


Deep unfolding provides a naturally suited framework for solving the CSOU sparse reconstruction problem. In CSOU applications, initial attempts adapted these frameworks with limited success. For instance, most existing unfolding methods, such as ISTA-Net \cite{zhang2018ista} and ADMM-Net \cite{yang2018admm}), employ static parameters. Their network weights and thresholds remain fixed, unable to adapt to input-specific characteristics. The dynamic unfolding paradigm emerged to address this. For instance, DISTA-Net \cite{han2025dista} generates convolutional weights adaptively, improving unmixing accuracy. Furthermore, SeqCSIST \cite{zhai2025seqcsist} further advanced this line by proposing a Temporal Deformable Feature Alignment module for enhanced unmixing performance in sequential data.



Our proposed DSCSNet shares the dynamic unfolding foundation with DISTA-Net \cite{han2025dista}. However, to directly address the core challenges of "powerful sparsity induction" and "fine-grained scene adaptivity" outlined in the introduction, DSCSNet introduces three key distinctions, constituting a significant extension and innovation within this dynamic framework:
\begin{enumerate}
    \item \textbf{Algorithmic Foundation:} DSCSNet is unfolded from the ADMM algorithm, not ISTA. ADMM's use of variable splitting decomposes complex optimization into simpler sub-problems, yielding a more stable process that better accommodates the "embedding of strong sparse constraints" required for CSOU.
    \item \textbf{Sparsification Core:} DSCSNet explicitly incorporates an L1-norm constraint, replacing the inherent proximal mapping sparsification of ISTA. This enhances sparsity induction, enabling more precise separation of blended spots into discrete targets.
    \item \textbf{Dynamic Mechanism:} DSCSNet employs an attention-based dynamic thresholding module. By perceiving local energy characteristics of the input spot, it generates pixel-wise dynamic thresholds. Compared to DISTA-Net's dynamic convolution weights, our approach allows for finer-grained, scene-adaptive adjustment—imposing strong sparsity on target regions and weak sparsity on the background.
\end{enumerate}

In summary, DSCSNet is the first CSOU method that integrates the ADMM framework with attention-based dynamic thresholding and a strong L1-norm sparsity constraint, effectively addressing the limitations of existing dynamic unfolding methods in sparsity induction and adaptive precision.

\section{Method} \label{sec:method}
\subsection{Problem Formulation} 
\label{sec:problemformulation}


\subsubsection{\textbf{Imaging model}}
Traditional imaging systems are constrained by the pixel size of the detector, which is directly related to the spatial sampling frequency.  This limitation restricts their ability to achieve sub-pixel-level target localization.  In contrast, CS takes advantage of the sparsity of the signal as a prior, effectively overcoming the constraints imposed by the Nyquist-Shannon Sampling Theorem (NSST) on spatial sampling frequency.  CS enables accurate scene reconstruction at sampling rates significantly lower than the Nyquist rate.  This unique capability allows sub-pixel localization to be reformulated as a sparse signal reconstruction problem.  Specifically, by discretizing the scene onto a high-resolution grid—finer than the resolution of the original image—the target is represented with greater sparsity in the grid space.  Leveraging this sparsity prior, we can model the long-range infrared small target imaging (CSIST) process as an underdetermined linear system based on the high-resolution grid.  By solving this system, we can achieve sparse signal reconstruction, enabling sub-pixel-level target localization.  This system consists of several components, which are described in detail as follows:

\begin{figure*}
    \centering
    \includegraphics[width=1\linewidth]{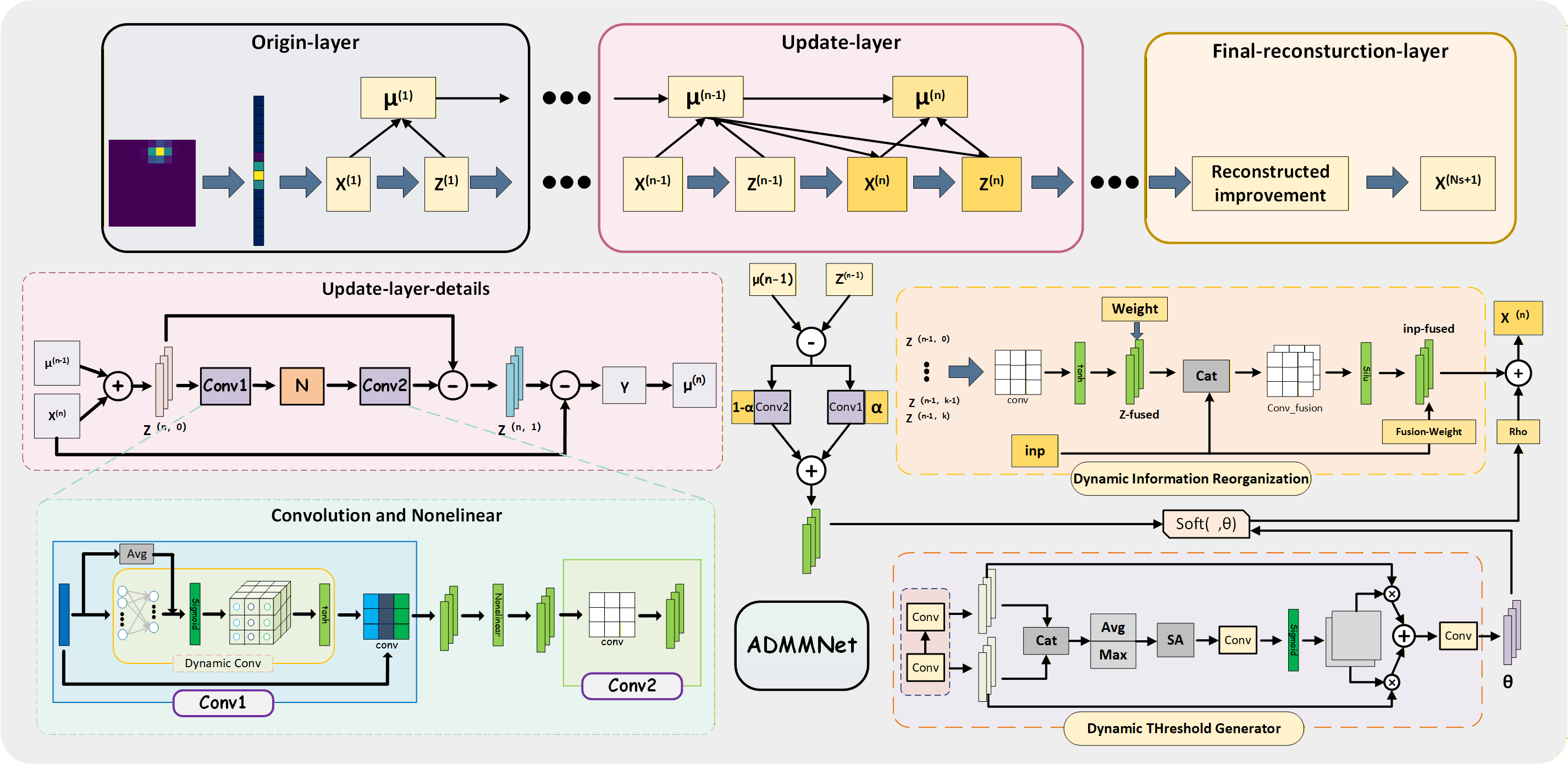}
    \caption{Architecture of the proposed Dynamic Convolutional Sparse Coding Network (DCSCNet). 
   }
    \label{fig:dcscnet_architecture}
\end{figure*}

\textit{\textbf{Sparse Target Vector}}: To achieve sub-pixel precision, the target scene is discretized into a high-resolution grid of size \( N_1 \times N_2 \) , where \(N_1\) and \(N_2\) represent the number of rows and columns of the grid, respectively. The target distribution flattened from this grid can be represented by a sparse vector \(\mathbf{x} = (x_1, x_2, \dots, x_N)\), which satisfies \( \mathbf{x} \in \mathbb{R}^N \) and \( \|\mathbf{x}\|_0 = K \ll N \), where \( \|\cdot\|_0 \) denotes the $\ell_0$-norm of a vector, \( N = N_1 N_2 \) represents the total number of grid points, \textit{K }represents the number of actual targets in the scene, and \textit{K }is much smaller than the total number of grid points \textit{N}.


\textit{\textbf{Point Spread Function (PSF)}}: Due to the diffraction effects of the optical system and the inherent physical limitations of the sensor, each point target on the high-resolution grid will not be represented as an ideal point on the actual low-resolution imaging plane. Instead, it will be blurred into a light spot with a specific distribution. This blurring process is typically modeled using the Gaussian point spread function (PSF):
\begin{equation}
    PSF(\textbf{q}; \textbf{p}_k) = \frac{1}{2\pi\sigma_{psf}^2} \exp\left(-\frac{||\textbf{q} - \textbf{p}_k||_2^2}{2\sigma_{psf}^2}\right) ,
\end{equation}
where $\textbf{q}=(q_{1},q_{2})$ denotes coordinates on the low-resolution imaging plane, and \( \|\cdot\|_2 \) denotes the $\ell_2$-norm. $\sigma_{psf}$ is the PSF width parameter, which is calibrated via optical system parameters or experimental data.


\textit{\textbf{Forward Imaging Model}}: In the actual remote infrared small target imaging scene, there is not only a single point target. After blurring the point spread function (PSF) of multiple point targets, their spots will be superimposed to form a blended spot. The blended spot is generated by the superposition of blurred infrared small targets, with the imaging process governed by physical constraints such as the optical diffraction limit, sensor resolution, and noise interference. The observed blended spot $y = (y_1, y_2, \dots, y_M) \in \mathbb{R}^{M}$, where M = \(M_1M_2\), generally denotes the number of pixels or the measurement dimension in the original observational data, and \(M_1\) and \(M_2\) represent the number of rows and columns of the grid of the original observational data, respectively. The resolution of this grid is much higher than the original observation resolution, typically satisfying \( N \gg M \). The observed  $\mathbf{y}$ is the linear superposition of blurred target spots and additive noise, modeled as:
\begin{equation}
    \mathbf{y} = \boldsymbol{\Phi} \cdot \mathcal{H}(\mathbf{x}) + \mathbf{n} ,
\end{equation}
where $\mathcal{H}$ denotes the convolution operation that models the blurring effect of the imaging system's point spread function (PSF), $\boldsymbol{\Phi}$ is the measurement matrix that performs a linear dimensionality reduction, projecting the high-dimensional blurred signal $\mathcal{H}(\mathbf{x})$ from the high-resolution grid into the low-dimensional observation space $\mathbb{R}^{M}$, and $\mathbf{n}$ is the addition of Gaussian noise.

\subsubsection{\textbf{Inverse Problem}}
CSOU aims to reconstruct the sparse, high-resolution signal \textbf{x} from the observed blended spot \textbf{y}, which is formulated as a constrained optimization problem:
\begin{equation}
\hat{\mathcal{H}(\mathbf{x})} = \arg\min_{\mathcal{H}(\mathbf{x})}\|{\Phi}\mathcal{H}(\mathbf{x}) - \mathbf{y}\|_2^2 +  \lambda \|\mathbf{x}\|_1,
\end{equation}
where \( \|\cdot\|_1 \) denotes the $\ell_1$-norm. Typical basis matrices $\mathcal{H}(\cdot)$  include the discrete cosine transform, discrete wavelet transform, and others. In this paper, we generalize this constrained standard CS model to a generalized form: let $\mathcal{H}(\mathbf{x}) = C\mathbf{x}$ where $C$ is an invertible matrix satisfying $C C^{-1} = I$. We then substitute $\mathcal{H}(\mathbf{x})$ with $C\mathbf{x}$ into the following model: 
\begin{equation}
    \hat{\mathbf{x}} = \arg\min_x \left\{ \frac{1}{2} \|\Phi \mathbf{x} - \mathbf{y}\|_2^2 + \sum_l \lambda_l |C_l \mathbf{x}| \right\},
    \label{compressed sense improve}
\end{equation}
where $C_l$ is the lth row of $C$. Since x is an image with arbitrary resolution, we further generalize the image representation $C_l \mathbf{x}$ from image convolution using fixed local filters and $\ell_1$ sparse regularization to a more general regularizer g(·). In Eq. \eqref{compressed sense improve}, we relax the requirements for orthogonality and invertibility of the basis matrix $C$, and thus we formulate the Generalized-CS model: 
\begin{equation}
    \hat{\mathbf{x}} = \arg \min_x \left\{ \frac{1}{2} \| \Phi \mathbf{x} - \mathbf{y} \|_2^2 + \sum_{l=1}^L \lambda_l g(D_l x) \right\},
\end{equation}
where $\textbf{D} = \{D_l\}$ denotes a transform matrix for a convolution operation, $L$ denotes the number of filters, and $\lambda = \{\lambda_l, \lambda\geq 0\}$ is the regularization parameter balancing data fidelity and sparsity. To enforce point-wise sparsity and facilitate adaptive learning, we introduce an auxiliary variable $\textbf{z}=\textbf{Dx}$, where $\mathbf{z} \in \mathbb{R}^{C}$ and $\textbf{D}\in \mathbb{R}^{C\times N}$ represent learnable sparse transformation matrices. The corresponding optimization problem is expressed through an augmented Lagrangian function:
\begin{small}
    \begin{equation}
\begin{aligned}
   \mathcal{L}_{\rho}(\mathbf{x}, \mathbf{z}, \mu) &= \frac{1}{2} \left\| \Phi \mathbf{x} - \mathbf{y} \right\|_2^2 + \sum_{l=1}^L \biggl[ \lambda_l g(z_l) \\
   &\qquad+ \langle \mu_l, D_l \mathbf{x} - z_l \rangle
+ \frac{\rho_l}{2} \left\| D_l \mathbf{x} - z_l \right\|_2^2 \biggr],\\
   s.t. z_l = D_l \mathbf{x}
\end{aligned}
\label{eq:augmented_lagrangian}
\end{equation}
\end{small}

where $\boldsymbol{\mu} = \{\mu_l\ ,\boldsymbol{\mu} \in \mathbb{R}^{C}\}$  , $\boldsymbol{\mu} \in \mathbb{R}^{C}$ is the Lagrange multiplier , and $\rho_l > 0$ is the penalty parameter controlling convergence stability. In Eq. \eqref{eq:augmented_lagrangian}, the first term denotes the data fidelity term, which enforces consistency between the reconstructed image and the actual observed data $\mathcal{y}$, $ \lambda_l g(z_l)$ represents sparsity-promoting regularization, the rest of this function consists of the constraint enforcement terms.\\
This problem exhibits a separable structure in the objective and a linear equality constraint, making it well‑suited for the ADMM algorithm. 
The ADMM algorithm solves this problem via alternating minimization steps, enhanced with dynamic threshold generation for adaptive sparse reconstruction. The specific iterative steps are as follows: 

\textbf{Step 1: x-update for Observation Consistency:} When $\mathbf{z}$ and $\boldsymbol{\beta}$ are fixed , the iteration for \textbf{x} proceeds as follows:
\begin{equation}
\begin{split}
    \mathbf{x}^{(k+1)} &= \arg\min_{\mathbf{x}^{(k)}} \frac{1}{2} \| \boldsymbol{\Phi}\mathbf{x}^{(k)} - \mathbf{y} \|_2^2 \\
    &+ \frac{\rho}{2} \left\| \mathbf{D}\mathbf{x}^{(k)} - \mathbf{z}^{(k)} + \frac{\boldsymbol{\mu}^{(k)}}{\rho} \right\|_2^2 ,
    \label{eq:x_update}
\end{split}
\end{equation}
where $k$ denotes the number of iterations. When $k$ = 0 , $\mathbf{x}$ represents the learnable network's input, and $\mathbf{z}$ and $\boldsymbol\beta$ are set to 0. This quadratic optimization admits a closed-form as follows:
\begin{equation}
 \mathbf{x}^{(k+1)} = (\boldsymbol{\Phi}^T \boldsymbol{\Phi} + \rho I) ^{(-1)}( {\Phi}\mathbf{x}^{(k)} - \rho ( \boldsymbol\beta^{(k)} - z^{(k)})) ,
 \label{original output}
\end{equation}
where $\mathbf{\beta}^{(k)}$ = $\mathbf{\frac{\boldsymbol{\mu}^{(k)}}{\rho}}$ and I indicates the identity matrix.

\textbf{Step 2: z-update for Sparsity Enforcement:} With $\mathbf{x}$ and $\boldsymbol{\mu}$ fixed, the iteration for $\mathbf{z}$ proceeds as follows:
\begin{equation}
    \mathbf{z}^{(k)} = \mathop{\arg\min}\limits_{{\mathbf{z}^{(k-1)}}}  \mathcal{L}(\mathbf{x}^{(k)}, \mathbf{z}^{(k-1)}, \boldsymbol{\mu}^{(k-1)}) .
    \label{z_larg_update}
\end{equation}

We address Eq. \eqref{z_larg_update} using the gradient descent method: 

\begin{equation}
\begin{split}
\mathbf{z^{(k)}}& =  \mathbf{z^{(k-1)}} - l_r \Bigg[ \rho \mathbf{z^{(k-1)}} - \rho \left( \mathbf{x^{(k)}} + \boldsymbol\beta^{(k-1)} \right) \\
           & + \sum_{l=1}^{L} \lambda_l D_l^T \mathcal{H}\left( D_l \mathbf{z^{(k-1)}} \right) \Bigg]\\
               &= \mu_1 \mathbf{z^{(k-1)}} + \mu_2 (\mathbf{x^{(k)}} + \boldsymbol\beta^{(k-1)}) - \\
               &\sum_{l=1}^{L} \tilde{\lambda}_l D_l^T \mathcal{H}(D_l \mathbf{z^{(k-1)}}) ,
\end{split}
\end{equation}

where  $\lambda_l$ denotes the weight of the regularization term for the $l$-th transform operator $D_l$ ,$D_l$ is the l-th sparse transform operator — typically implemented as a learnable convolutional filter bank , mapping the intermediate variable $\mathbf{z}$ into a feature space where the representation is encouraged to be sparse , $\mu_1 = 1 - l_r\rho$ , $\mu_2 = l_r\rho$ and $\tilde\lambda_l = l_r\lambda$. The subtraction part's operation can be split into two convolutions $C_1(\cdot)$, $C_2(\cdot)$ and one activation  $\mathcal{H(\cdot)}$. 

\textbf{Step3: $\beta$-update for Constraint Enforcement:}
Update the Lagrange multiplier to enforce the constraint $\mathbf{z} = \mathbf{D}\mathbf{x}$:
\begin{equation}
    \boldsymbol{\beta}^{(k+1)} = \boldsymbol{\beta}^{(k)} + \rho \left( \mathbf{x}^{(k+1)} - \mathbf{z}^{(k+1)} \right).
    \label{eq:mu_update}
\end{equation}

The integration of learnable components—sparse transform $\mathbf{D}$, and adaptive coefficients $(\lambda, \rho)$—enables the framework to adapt to dynamic target characteristics, learn target-specific sparse representations beyond fixed priors, and achieve precise separation of overlapping targets via $\ell_1$-norm-induced sparsity with enhanced adaptability. This dynamic ADMM optimization framework forms the mathematical foundation of DSCSNet, where each iteration is unfolded into a network layer for end-to-end training.

\subsection{Dynamic Sparse Compressing Sensing Net} \label{sec:DSCSNet}

DSCS-Net integrates the dynamic parameter generation with the ADMM unrolling framework of ADMM-CSNet. The network comprises $N$ cascaded stages, each corresponding to one ADMM iteration. As illustrated in \textbf{Fig. 3}, each stage $k \in \{1,2,...,N \}$ contains three core components: \textbf{Dynamic Transform \& Threshold Module} $\mathcal{F}^{(k)}$: Learns input-adaptive feature representations, $\Theta_d^{(k)}$: Generates data-dependent shrinkage parameters; \textbf{Reconstruction Module}: Recovers sparse target signals through linear projection.

\subsubsection{\textbf{Dynamic Transform with L1 Regularization}}
To meet the requirements for extreme sparsity during the iterative process, as well as to ensure the robustness of the network's iterative enhancement during the iterative stage. We embed $\ell_1$ regularization into dynamic convolution to achieve adaptive iteration based on input content, as shown in Eq. \eqref{z_argmin}:

\begin{equation}
\begin{split}
    \mathbf{z^{(k)}} &= \arg \min_{\mathbf{z}^{(k)}} { \frac{\rho_l}{2} \|D_l \mathbf{x^{(k)}} + \boldsymbol\beta_l^{(k-1)} - \mathbf{z}\|_2^2 + \lambda_l \|\mathbf{z}\|_1 }  .
    \label{z_argmin}
\end{split}
\end{equation}

In Eq.8, our solution employs a dynamic convolution  for adaptive feature extraction, The module $\mathcal{F}^{(k)}$ replaces traditional fixed transforms with learnable sparse representations:
\begin{equation}
\begin{split}
   \mathcal{F}^{(k)}(\mathbf{z^{(k)}}) = C_2 (\text{SiLu}(C_1(W, \mathbf{z^{(k)}}))) ,
\end{split}
\end{equation}
where  $W = f(\cdot)$ generates dynamic convolutional weights from inputs, $C_1(\cdot)$ represents using the dynamic convolutional weights for convolutional operation and $C_2(\cdot)$ represents using the static convolutional weights for convolutional operation.

\subsubsection{\textbf{Sparsity and robustness enhancement in the final reconstruction stage }}
The original algorithm uses Eq. \eqref{original output} as the final reconstruction output.  However, due to differences between this task and traditional compressed sensing, this approach yields low image quality and poor restoration accuracy.  In order to overcome the problem that the original ADMM depth expansion algorithm cannot adapt to the sparsity of the CSOU task and the robustness is reduced in the face of scenes with a large number of small targets, we add Dynamic Threshold Generator (DTG) and Dynamic Information Reorganization (DIR) in the reconstruction stage. The final reconstruction stage is formulated as Eq. \eqref{final output}:

\begin{equation} 
\begin{split}
\mathbf{x}^{(k+1)} &= \left({\Phi}^T {\Phi} + \rho I\right) ^{(-1)}\\
&\cdot\left ( {f_{\text{DIR}}(\Phi}\mathbf{x}) -\rho \cdot \text{Soft}(f_{\text{DTG}}( ( \delta^{(k)}))\right),
\label{final output}
\end{split}
\end{equation}
where $\delta^{(k)}=\beta^{(k)} - \mathbf{z}^{(k)}$, $f_{\text{DTG}}(\cdot)$ employs spatial attention to capture target density variations, $f_{\text{DIR}}(\cdot)$ integrates the original input with iterative information while simultaneously enhancing sparsity, and $\text{Soft}(\cdot) = \text{sign} (\cdot )  \cdot \max(0, |\cdot| - \theta)$ denotes soft-thresholding operation.

    \textbf{(a) Dynamic Threshold Generator Module:} 
In the output of the original reconstruction stage, for $\delta^{(k)}$, it made full use of the iterative information and we believe that it can meet the requirements in terms of sparsity. However, in scenarios where the number of small targets increases, the network has difficulty predicting a specific small target precisely to a particular pixel, resulting in varying degrees of ghosting in adjacent pixels. Therefore, we believe that in the final reconstruction stage, we need a threshold that can be generated based on the input content to finely process $\delta^{(k)}$, reducing the ghosting caused by the increase in the number of small targets, thereby enhancing the overall robustness of the network. This module employs spatial attention to perceive local energy distributions and generates pixel-wise adaptive thresholds, lowering values in target regions to preserve weak signals and raising them in background areas to suppress noise, thereby enabling content-aware filtering. 

The DTG processes multi-scale features through parallel convolutions and spatial attention:
\begin{align}
    \tilde{U} &= [\text{Conv}_{3\times3}^{(1)}(\mathcal{F}_d^{(k)}), \text{Conv}_{3\times3}^{(2)}(\mathcal{F}_d^{(k)})] ,\\
    \theta_d^{(k)} &= \text{Conv}\left(\left[\text{AvgPool}(\tilde{U}); \text{MaxPool}(\tilde{U})\right]\right).
\end{align}

The adaptive soft-thresholding operation becomes:
\begin{equation}
   \hat{\delta^{(k)}} = \text{Soft}(\delta^{(k)}) = \text{sign} ({\delta^{(k)}} )  \cdot \max(0, |\delta| - \theta_d^{(k)}),
    \label{eq:adaptive_soft_threshold}
\end{equation}
where $\hat{\delta^{(k)}}$ means the $\delta^{(k)}$ after being processed by the soft threshold operation. The threshold parameter $\theta_d^{(k)}$ is adaptively generated through a dedicated Dynamic Threshold Generator (DTG) module:
\begin{equation}
    \theta_d^{(k)} = f_{\text{DTG}}\left(\mathcal{F}_d(\delta^{(k)})\right),
\end{equation}
where $\mathcal{F}_d(\cdot)$ uses static convolution and dynamic convolution to extract multi-scale features.

\textbf{(b)Dynamic Information Reorganization Module:}
In the ADMM algorithm, the auxiliary variable $\mathbf{z}$ from early iterations is typically discarded. As shown in Eq. \eqref{original output}, this discarding leads to insufficient sparse representation ability in the final reconstruction layer, particularly for the initialized image input part, due to the lack of iterative sparsity refinement and the difficulty of enhancing sparsity via content-based adaptive filtering. This inherent mismatch between the network structure and the task requirements limits performance in scenarios like CSOU. To address this, we design a Dynamic Information Reorganization module (DIR) that aggregates auxiliary variables from the past $m$ iterations. These historical features are integrated and reintroduced into the current reconstruction via residual connections, enhancing the consistency of sparse estimation across layers and suppressing artifacts while maintaining high interpretability. The auxiliary variable $\mathbf{z}$ involved in the fusion is: 

\begin{equation}
    \mathbf{Z}^{(i)} = \bigl\{\mathbf{z^{(i)}}\bigr\}_{i=n-m}^{n},
\end{equation}
Let $m$ be the number of auxiliary variables $\mathbf{z}$ fused from $n$ iterations (e.g., last $m$ if $m=3$, $n=6$). After sparsifying each $\mathbf{z}^{(i)}$ to $\mathbf{Z}_{\text{enhanced}}^{(i)}$, they are averaged.
\begin{equation}
    \text{Z}_{\text{enhanced}}^{(i)} = \tanh(\text{Conv}(\text{Z}^{(i)})),
\end{equation}
\begin{equation}
\mathbf{z}_{\text{fused}} = \text{Weight}(\text{Z}_{\text{enhanced}}^{(i)}).
\end{equation}

Finally, the sparsity-enhanced initialized input features are fused with the enhanced auxiliary variable $\mathbf{z}$:

\begin{equation}
  inp_{enhanced} = \text{Conv}( \text{SiLU}(\text{Conv} (\text{Cat} [\mathbf{z}_{\text{fused}},\mathcal{F}(inp)])),
\end{equation}
where $\mathcal{F}$ means the sparsity-enhanced initialization, $inp$ denotes oringinal input, $inp_{enhanced}$ represents sparsity-enhanced input .\\
In summary, as illustrated in \textbf{Fig.~4}, the network comprises three stages: Origin-layer, Update-layer, and Final-reconstruction-layer. The Origin-layer encodes the input into initial features $X^{(1)}$, sparse codes $Z^{(1)}$, and multipliers $\mu^{(1)}$. The Update-layer iteratively refines $X^{(n)}$, $Z^{(n)}$, and $\mu^{(n)}$ via convolutional proximal operators (Conv1, Conv2) with nonlinear transforms and dynamic residual learning, fusing historical $Z^{(n,0)}$ and current $Z^{(n,1)}$ to update sparse codes. Within this stage, the Dynamic Information Reorganization module aggregates historical sparse code sequences ${Z^{(n-1,k)}}$ through convolutional fusion and adaptive weighting, producing updated $X^{(n)}$; the Dynamic Threshold Generator adaptively predicts soft thresholds $\theta$ from current sparse states via channel attention, while the coefficient $\rho$ enables iterative threshold optimization. Finally, the Final-reconstruction-layer outputs $X^{(N_s+1)}$.

\subsection{Loss Function for Sparse Recovery}
The training objective combines reconstruction fidelity:
\begin{equation}
    \mathcal{L}_{\text{discrepancy}} = \frac{1}{MN_s}\sum_{i=1}^M \|\tilde{s}_i^{(N)} - s_i\|_2^2.
\end{equation}
where $\mathcal{L}_{\text{discrepancy}}$ is the mean squared error between the reconstructed signal $s_i^{(N)}$ and the ground truth $s_i$.



\section{Experiments} \label{sec:experiment}

\subsection{Experimental Settings} \label{subsec:setting}
\begin{figure}    
\centering    
\includegraphics[width=0.45\textwidth]{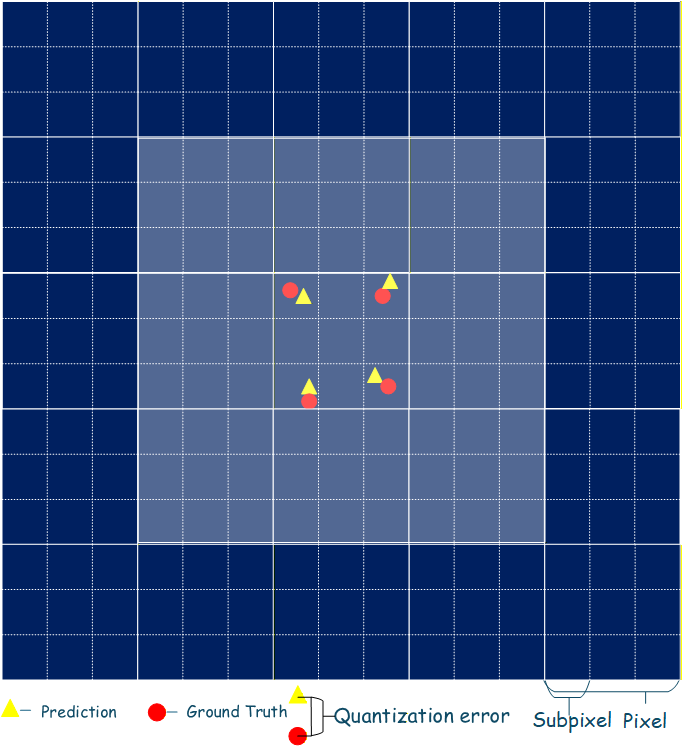}    
\caption{Each pixel is subdivided into an 3 × 3 sub-pixel grid that represents the set of potential target locations.}    
\label{fig:CSOmetric}
\end{figure}
\subsubsection{\textbf{Evaluation Metrics}}
\label{subsubsec:evaluationmetrics}
\begin{table*}[htbp]
  \renewcommand\arraystretch{1.4}
  \footnotesize
  \centering
  \caption{Systematic Benchmarking Analysis of DCSCNet Performance: Comparative Evaluation Against Multiple Method Categories Including Classical Optimization, Super-Resolution Networks, and Deep Unfolding Architectures on \textbf{CSIST100K} Dataset.}
  \label{table:Comparison with State-of-the-Arts methods on SeqCSIST dataset}
  \vspace{-2pt}
  \setlength{\tabcolsep}{6.pt}
  \begin{tabular}{l|cc|cccccc}
  \multirow{2}{*}{\textbf{Method}}    & \multirow{2}{*}{\textbf{Params} $\downarrow$} & \multirow{2}{*}{\textbf{FLOPs} $\downarrow$} & \multicolumn{6}{c}{CSO-mAP $\uparrow$}   \\
  & & &  {\textbf{CSO-mAP}} & \textbf{AP$_{05}$} & \textbf{AP$_{10}$} & \textbf{AP$_{15}$} & \textbf{AP$_{20}$} & \textbf{AP$_{25}$} \\
  \Xhline{1pt}
  \multicolumn{9}{l}{\fontsize{10.5}{13}\selectfont \textit{Traditional Optimization}} \\ \hline
  ISTA~\cite{Daubechies2004ISTA} &  & & 7.46  & 0.01  & 0.31  & 2.39  & 9.46 & 25.14 \\
  \hline
  \multicolumn{9}{l}{\fontsize{10.5}{13}\selectfont \textit{Image Super-Resolution}}  \\
  \hline
  SRCNN~\cite{dong2015image}  & 0.019 M & 1.345 G & 29.06 & 0.23 & 4.10 & 21.65 & 49.95 & 69.39 \\
  \hline
  ACTNet~\cite{zhangACTNet2023}  & 46.323 M & 62.80 G & 45.61 & 0.38 & 7.46 & 41.13 & 83.12 & 95.95 \\
  \hline
  CTNet~\cite{lim2017enhanced}  & 0.400 M & 2.756 G & 45.11 & 0.38 & 7.53 & 40.39 & 82.11 & 95.14 \\
  \hline
  DCTLISA~\cite{zeng2023}  & 0.865 M & 13.69 G & 44.51 & 0.39 & 7.35 & 39.35 & 81.15 & 94.34 \\
  \hline
  SRFBN~\cite{li2019gated} & 0.373 M & 3.217 G & 46.05 & 0.43 & 8.31 & 42.83 & 83.72 & 94.95 \\
  \hline
  RDN~\cite{zhang2018residual} & 22.31 M & 173 G & 45.81 & 0.35 & 7.11 & 41.07 & 84.07 & 96.43 \\
  \hline
  EDSR~\cite{lim2017enhanced} & 1.552 M & 12.04 G & 45.32 & 0.33 & 7.07 & 40.58 & 83.24 & 95.41 \\
  \hline
  HIT-SIR~\cite{zhang2024unleashing} & 0.769 M & 7.357G & 44.51 & 0.40 & 7.16 & 39.55 & 80.94 & 94.50 \\
  \hline
  HiT-SNG~\cite{zhang2024unleashing}  & 0.952M & 13.324G & 45.01 & 0.39 & 7.34 & 40.19 & 81.98 & 95.17 \\
  \hline
  HiT-SRF~\cite{zhang2024unleashing} & 0.844M & 7.945G & 44.71 & 0.40 & 7.71 & 40.03 & 80.94 & 94.49 \\
  \hline
  \multicolumn{9}{l}{\fontsize{10.5}{13}\selectfont \textit{Deep Unfolding}}  \\
  \hline
  LIHT~\cite{wang2016learning}  & 21.10 M & 1.358 G & 10.35 & 0.06 & 0.92 & 4.99 & 14.74 & 30.50 \\
  \hline
  LAMP~\cite{metzler2017learned}  & 2.13 M &0.278 G & 14.22 & 0.05 & 1.11 & 7.31 & 21.56 & 41.06 \\
  \hline
  ISTA-Net~\cite{zhang2018ista} & 0.171 M & 12.77 G & 45.16 & 0.41 & 7.71 & 40.57 & 82.58 & 94.53 \\
  \hline
  ISTA-Net+~\cite{zhang2018ista} & 0.38 M & 7.70 G & 51.02 & 1.00 & 13.70 & 52.70 & 90.40 & 93.70 \\
  \hline
  ISTA-Net++~\cite{you2021ista}  & 0.337 M & 24.33 G & 46.06 & 0.42 & 7.66 & 41.58 & 84.46 & 96.17 \\
  \hline
  LISTA~\cite{gregor2010learning} & 21.10 M & 1.358 G & 30.13 & 0.25 & 4.13 & 22.29 & 51.18 & 72.80 \\
  \hline
  FISTA-Net~\cite{Xiang2021fista} & 0.074 M & 18.96 G & 44.66 & 0.41 & 7.58 & 39.74 & 81.24 & 94.19 \\
  \hline
  TiLISTA~\cite{liu2019alista}  & 2.126 M & 0.278G & 14.95 & 0.06 & 1.23 & 7.72 & 22.50 & 46.23 \\
  \hline
  DISTANet~\cite{han2025dista}  & 2.179 M & 35.10 G & 45.94 & 0.38 & 7.54 & 41.26 & 84.03 & 96.35 \\
  \hline
  \rowcolor[rgb]{0.9,0.9,0.9} $\star$ \textbf{DCSCNet (Ours)}  & 6.23M & 40.70 G & \textbf{\underline{46.36}} & 0.44 & 7.93 & 41.50 & 85.40 & 96.54 \\
  \end{tabular}
  \vspace{-1\baselineskip}
\end{table*}

\begin{figure*}    
\centering    
\includegraphics[width=1\textwidth]{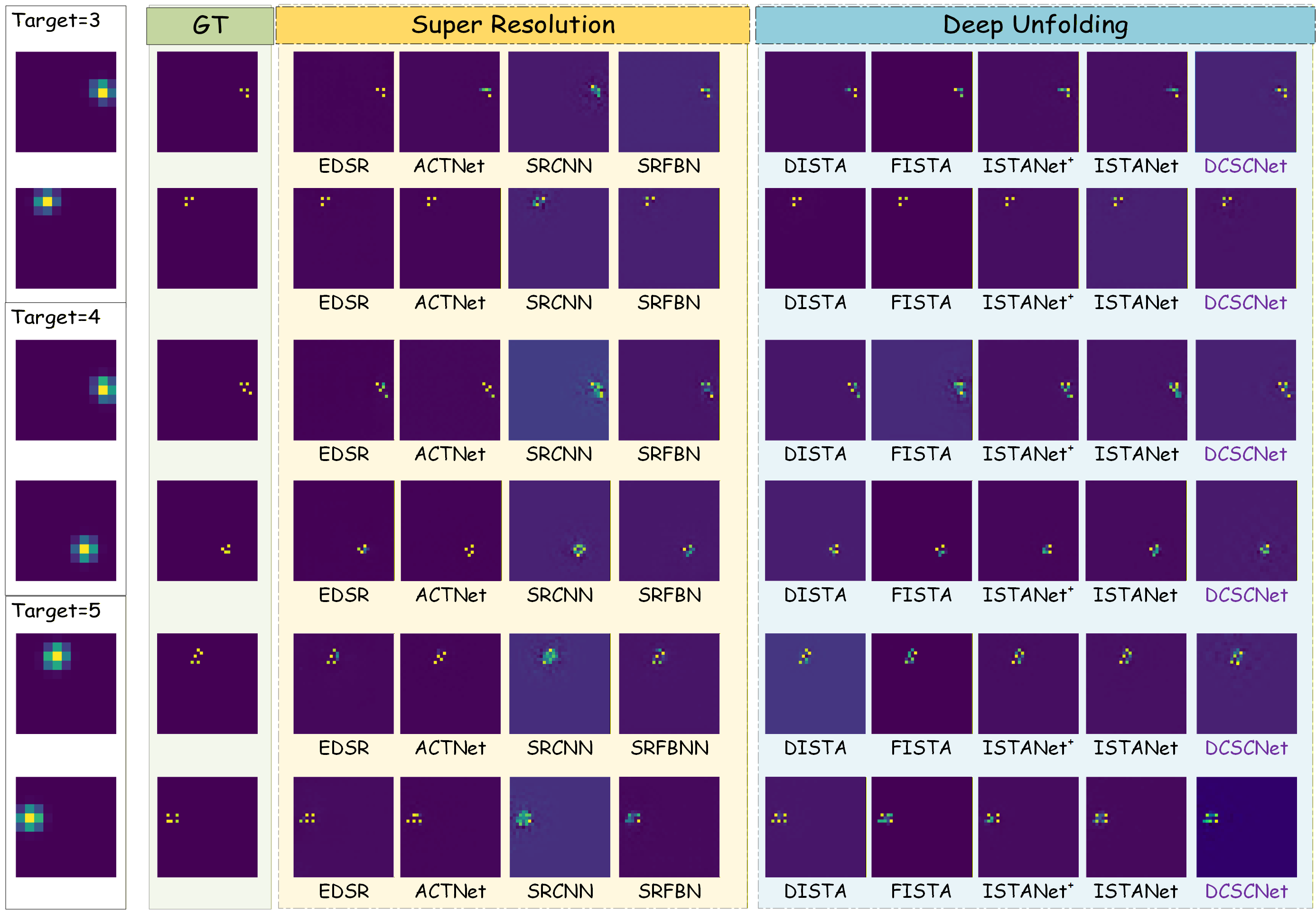}     
\caption{Visual comparison of 3× sub-pixel partition reconstruction in scenes with varying numbers of adjacent targets}
\label{subsubsec:visualization}
\end{figure*}
To evaluate DCSCNet, we adopted the CSO mean Average Precision (CSO-mAP) as the metric to assess network performance, a metric specifically designed for the CSOU task.

The evaluation process is as follows: First, each prediction is classified as True-Positive (TP) or False-Positive (FP); then, a binary list is created. In this binary list, TP predictions are encoded as 1 and FP predictions are denoted by 0. This binary encoding underpins the precision–recall (PR) curve. The confidence threshold $\delta$ for positive predictions is then systematically varied to obtain multiple pairs of Precision and Recall values. These values collectively trace the precision–recall curve. Average Precision (AP) is defined as the area under the PR curve, which offers a fine-grained assessment of model performance across all confidence levels and jointly accounts for the precision–recall trade-off. In this task, we set the confidence threshold $\delta$ to the values {5, 10, 15, 20, 25}, which indicates the indicators $AP_{05},  AP_{10}, AP_{15}, AP_{20}, AP_{25}$, and on this basis, we computed the network-performance metric CSO-mAP.

\label{subsubsec:trainingsettings}

We implemented DCSCNet on the open-source MMCV framework, integrating and adapting existing compressed-sensing techniques for the hyperspectral unmixing of infrared small-target imagery.

\textbf{Train Settings:}
As a downstream stage of infrared small-target detection, DSCSNet receives target patches cropped from the detection results rather than the original full image.We used images from the CSIST-100K dataset as input, where the target patches are 11 × 11 pixels. The network outputs the predicted target position, sub-pixel offset, and radiometric intensity. Throughout the experiment, the initialized super-resolution ratio \textit{c}is set to 3, and both the ground-truth and the unmixed output images have a spatial dimension of 33 × 33 pixels. The initialization parameter c is introduced to subdivide each pixel into a grid of n × n, producing units of sub-pixel $L = UV n^{2}$. Provided the grid is sufficiently dense, each sub-pixel cell contains at most one target object, and the upper bound resulting in the localization error is $\frac{\sqrt{2}}{2n}D$.Specifically, for the i-th target, its position and radiance are denoted by $(x_i, y_i, s_i)$. After initialization, the radiance value $s_i$ is assigned to the pixel whose coordinates are $( c \cdot x_i + \frac{c - 1}{2}, \; c \cdot y_i + \frac{c - 1}{2} )$. Remaining pixels are set to zero.By adjusting c, we guaranty that each sub-pixel grid contains at most one target. This allows iterative updates to refine the sub-pixel position and accurately recover the target’s radiance. We trained the network with the Adam optimizer under the MMEngine framework. The learning rate was fixed at $1 × 10^{-4}$.

\textbf{Test Settings:}
To quantify the sub-pixel localization accuracy of DCSCNet on the infrared small-target unmixing task, we measured each model’s performance via CSO-mAP. We extracted bright pixels whose intensity exceeded a threshold of 50 from the unmixed image and treated them as predicted targets. Their intensity values were recorded as peak strengths. Their positions were re-projected into the original 11 × 11 coordinate system. The re-projection formulas are $x'=\left\lfloor\dfrac{x-\bigl\lfloor\frac{c-1}{2}\bigr\rfloor}{c}\right\rfloor,\quad$$y'=\left\lfloor\dfrac{y-\bigl\lfloor\frac{c-1}{2}\bigr\rfloor}{c}\right\rfloor$.

\subsection{Comparison with State-of-the-Arts} \label{subsec:sota}

\subsubsection{Quantitative Analysis}
\label{subsubsec:quantitative}

Table~\uppercase\expandafter{\romannumeral1} compares DCSCNet with three method families on CSIST-100K: classical optimization (ISTA), super-resolution networks (SR), and deep-unfolding networks (ACTNet, ISTANet). Evaluation covers parameters, FLOPs, and CSO-mAP.
\begin{enumerate}
\item \textbf{Compared with traditional model-driven methods (e.g., ISTA):} DCSCNet achieves 46.36\% mAP versus 7.46\%, revealing the limitations of purely model-driven approaches. By integrating model- and data-driven strategies, it effectively handles overlapping targets and scene variations, yielding substantially higher unmixing and localization accuracy.
\item \textbf{Compared with super-resolution methods:} SR methods perform poorly in CSOU. DCSCNet incorporates physics-matched dynamic thresholds and sparse iterations to capture finer details under extreme sparsity, conducting reconstruction early in the network rather than at the final stage, enabling more effective handling of complex target scenarios.
\item \textbf{Compared with deep unfolding methods:} DCSCNet dynamically integrates iterative information, excelling in both lenient ($AP_{25}$: 96.54\%) and ultra-strict ($AP_{05}$, $AP_{10}$) metrics. While ADMMCSNet achieves reasonable $AP_{25}$ with deeper iterations, it lags in stringent localization. DCSCNet ranks first overall among deep-unfolding methods, maintaining robust performance across thresholds.
\end{enumerate}
    The CSO-mAP metric is adopted with $AP_{05}$–$AP_{25}$ reported over a 0.05–0.5 pixel range (step 0.05). Given the systematic error of 0.236 pixels from the $c=3$ grid, $AP_{20}$ and $AP_{25}$ serve as primary metrics, where DCSCNet achieves 85.40\% and 96.54\%, respectively—on par with DISTANet—while leading in sub-pixel localization ($AP_{05}$, $AP_{10}$). In dense multi-target scenarios, DCSCNet attains 46.36\% CSO-mAP, confirming robustness across precision levels. These results are achieved with only 6.23M parameters and 40.7G FLOPs, yielding over $7\times$ parameter compression and at least $1.5\times$ FLOPs reduction versus ACTNet and HAN, significantly easing deployment on embedded platforms.

\begin{table*}[htbp]
\caption{Quantitative ablation results on the CSIST 100K dataset. We report the Average Precision (AP) averaged over different thresholds, along with the primary evaluation metric CSO-mAP.}
\centering
\resizebox{\textwidth}{!}{%
\begin{tabular}{c|ccc|cccccc}
\toprule
\multirow{2}{*}{\textbf{Strategy}} & \multicolumn{3}{c|}{\textbf{Module}} & \multicolumn{6}{c}{\textbf{CSIST 100K}} \\
& DCN & DIR & DTG& mAP & $AP_{25}$& $AP_{20}$ & $AP_{15}$ & $AP_{10}$ & $AP_{05}$ \\
\midrule
DCSCNet & \ding{51} & \ding{51} & \ding{51} & 46.36 & 0.44 & 7.93 & 41.50 & 85.40 & 96.54 \\
ADMMCSNet & \ding{55} & \ding{55} & \ding{55} & 43.64 & 0.40 & 7.93 & 38.80 & 79.27 & 91.80 \\
DCSCNet w/o DCN & \ding{51} & \ding{55} & \ding{55} & 44.90 & 0.41 & 7.42 & 39.98 & 83.32 & 93.44 \\
DCSCNet w/o DIR & \ding{51} & \ding{51} & \ding{55} & 44.35 & 0.40 & 7.51 & 40.37 & 83.85 & 94.61 \\
DCSCNet w/o DTG & \ding{51} & \ding{55} & \ding{51} & 45.94 & 0.39 & 7.62 & 41.31 & 84.60 & 95.80 \\
\bottomrule
\end{tabular}%
}
\end{table*}

\subsubsection{Visualization Analysis}To more clearly demonstrate the advantages of the proposed model, partial experimental visualization results are provided, as shown in \textbf{Fig. 6}. Specifically, \textbf{Fig. 6} presents comparative reconstruction experiments under the 3× sub-pixel partitioning framework, focusing on typical scenes containing 3–5 spatially adjacent infrared small targets. This evaluation is designed to assess the analytical performance of each algorithm under high target overlap conditions.
    \begin{enumerate} 
        \item \textbf{Compared with traditional model-driven methods}: Traditional methods and existing deep unfolding networks exhibit varying degrees of boundary blurring and target merging. Moreover, as the number of targets increases to five, both localization errors and miss detection rates rise significantly.
        \item \textbf{Compared with super-resolution methods}: Super-resolution methods are inconsistent with the physical principles of the CSOU task. They rely heavily on the contribution of the initialization method to the task. Unlike optimization-based networks, if the initialization is insufficiently accurate, such methods cannot iteratively refine the image to improve metrics. Additionally, the localization accuracy and radiation intensity restoration of super-resolution methods are highly unstable.
        \item \textbf{Compared to deep unfolding methods}: As the number of targets increases, existing deep unfolding methods exhibit varying degrees of ghosting artifacts and introduce noticeable errors in radiation intensity restoration. In contrast, DCSCNet, by leveraging the synergistic enhancement mechanism of iterative information-based dynamic reconstruction and adaptive thresholding, can simultaneously maintain target count consistency, edge clarity, and spatial distribution accuracy at the sub-pixel level. This validates its robustness in dense multi-target unmixing tasks.
    \end{enumerate}
\subsection{Ablation Study} \label{subsec:ablation}
\subsubsection{Impact of different components}
We conduct ablation studies to evaluate each component's contribution, with results in TABLE~\uppercase\expandafter{\romannumeral2}.

"DCSCNet w/o DCN" removes dynamic information reorganization and soft thresholding from ADMMCSNet, retaining only static convolution. Adding DCN enables sparse extraction across iterations, improving all metrics with minimal cost (e.g., $AP_{25}$ from 91.80\% to 93.44\%), confirming its suitability.

"DCSCNet w/o DIR" removes dynamic information reorganization. Compared to "w/o DCN", $AP_{25}$ rises to 94.61\%, showing that integrating iterative information reduces loss and enhances final sparse representation.

"DCSCNet w/o Thres" removes dynamic soft thresholding. $AP_{25}$ increases to 95.80\% versus "w/o DCN", indicating that input-adaptive thresholding suppresses noise, aligns with physical priors, and improves interpretability and performance.

Full DCSCNet outperforms baseline ADMMCSNet: CSO-mAP from 45.16\% to 46.36\%, $AP_{25}$ from 91.80\% to 96.54\%, validating synergistic gains. Removing DIR drops CSO-mAP to 44.35\% and $AP_{25}$ to 94.61\%, confirming its role in adaptive feature correction. Removing Thres reduces CSO-mAP to 45.94\% and $AP_{25}$ to 95.80\%, highlighting its necessity for artifact suppression and localization accuracy.
\begin{figure}    
\centering    
\includegraphics[width=0.8\linewidth]{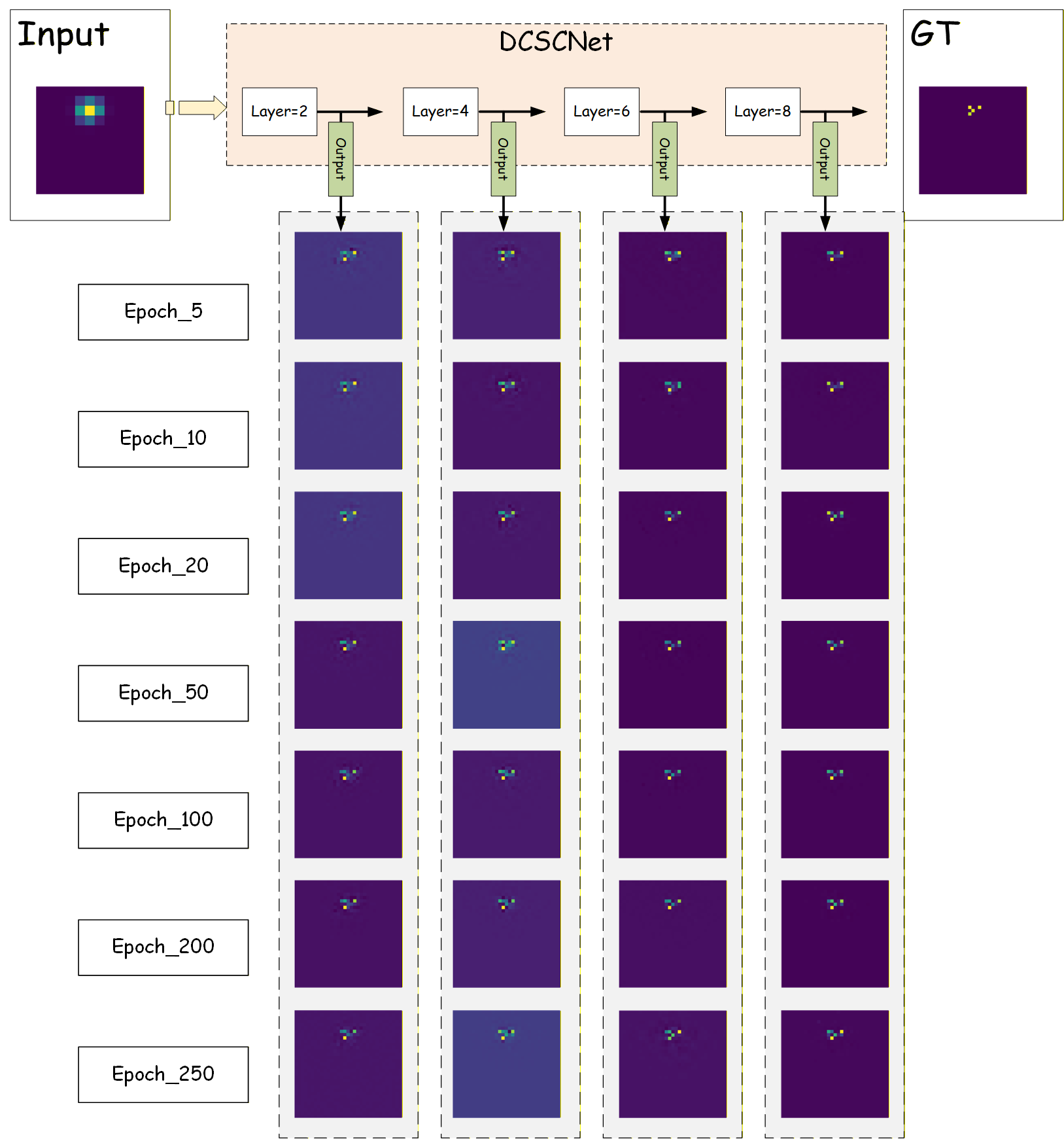}    
\caption{The degree of unmixing varies with the number of iteration rounds and training epochs}    
\label{fig:CSOmetric}
\end{figure}
\begin{figure}    
\centering    
\includegraphics[width=0.5\textwidth]{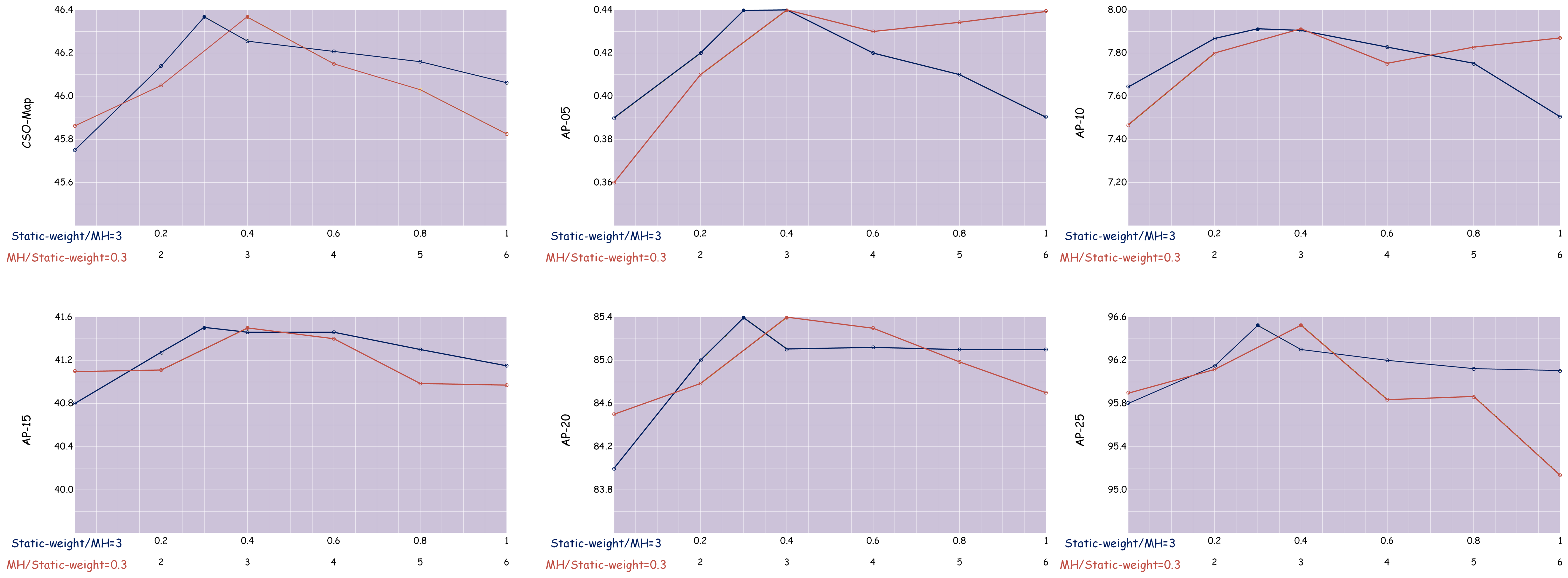}
\caption{Hyperparameter analysis}   
\label{fig:CSOmetric}
\end{figure}
\subsubsection{Hyperparameter analysis}

    We trained networks with layer counts 2, 4, 6, 8, 10. Performance peaked at 6 layers (46.36\% CSO-mAP, as shown in \textbf{Fig.~7}), with best reconstruction at 6/250 epochs and 8/200 epochs. Six layers provide the optimal accuracy–efficiency trade-off.\\
    \textbf{(a) Dynamic weight:} varied from 0 to 1; best at 0.7, stable 0.5-0.9, drop at 1 (45.3\%), confirming coefficient importance.\\
    \textbf{(b) Dynamic reorganization layer:} with 6 layers, tested positions 1–6; best at 3; beyond this, model complexity increases and convergence speed decreases, yielding marginal gains; optimal at 3.


\section{Conclusion} \label{sec:conclusion}
This paper addresses the ill-posed CSOU problem, where existing methods struggle to balance sparsity and adaptability. To resolve this, we propose DSCSNet, coupling ADMM with deep unfolding. Key innovations: strict $\ell_1$-norm sparsity in ADMM updates to avoid smoothing bias, plus self-attention-based dynamic threshold and reorganized sparse iteration mechanisms for adaptive sparsification. The model preserves compressed sensing physics while ensuring robust adaptability. Experimental results show DSCSNet outperforms SOTA in accuracy and generalization. Future work includes multi-modal extension and real-time optimization.

\bibliographystyle{IEEEtran}
\bibliography{reference.bib}

\end{document}